\documentclass[utf8]{FrontiersinHarvard} 

\usepackage{url}
\usepackage{lineno}
\usepackage[onehalfspacing]{setspace}

\usepackage[
per-mode=fraction,
separate-uncertainty=true,
multi-part-units=single
]{siunitx}
\usepackage{comment}
\usepackage[normalem]{ulem}
\usepackage{xspace}
\usepackage{xcolor}

\DeclareSIUnit\px{px}
\DeclareSIUnit\fps{fps}

\definecolor{OliveGreen}{RGB}{0,200,25}

\newcommand{\ie}{i.\,e.\xspace}
\newcommand{\eg}{e.\,g.\xspace}

\newcommand{\armarVI}{\mbox{ARMAR-6}\xspace}

\newcommand{\armarx}{\mbox{ArmarX}\xspace}

\newcommand{\ackJuBotOML}{This work has been supported by the Carl Zeiss Foundation through the JuBot project and by the German Federal Ministry of Education and Research (BMBF) through the OML project (01IS18040A).}

\usepackage{amsmath}
\usepackage{amssymb}

\usepackage{multicol}
\usepackage{rotating}
\usepackage{tabularx}
\usepackage{booktabs}
\usepackage{array}

\usepackage{mathtools}
\usepackage{nicefrac}
\usepackage{relsize}
\usepackage{xparse}
\usepackage{subcaption}
\usepackage{makecell}
\usepackage[export]{adjustbox}

\usepackage{hyperref}
\usepackage[capitalize]{cleveref}  

\linenumbers

\def\keyFont{\fontsize{8}{11}\helveticabold }
\def\firstAuthorLast{Kartmann and Asfour} 

\def\Authors{Rainer Kartmann\,$^{*}$ and Tamim Asfour}

\def\Address{
  The authors are with the 
  High Performance Humanoid Technologies Lab, 
  Institute for Anthropomatics and Robotics, 
  Department of Informatics, 
  Karlsruhe Institute of Technology (KIT), 
  Karlsruhe, 
  Germany
}

\def\corrAuthor{Rainer Kartmann}

\def\corrEmail{rainer.kartmann@kit.edu}

\crefname{section}{Section}{Sections}
\crefname{figure}{Fig.}{Figs.}
\crefname{table}{Table}{Tables}
\crefname{equation}{}{}
\Crefname{equation}{Equation}{Equations}

\newcommand{\reals}{\mathbb{R}}

\newcommand{\SOthree}{\mathrm{SO(3)}}
\newcommand{\SEthree}{\mathrm{SE(3)}}

\newcommand{\styleset}[1]{\mathcal{#1}}
\newcommand{\mvec}[1]{\boldsymbol{\mathbf{#1}}}
\newcommand{\mset}[1]{\left\{ #1 \right\}}
\newcommand{\mtup}[1]{\left( #1 \right)}

\newcommand{\angl}{\phi}

\newcommand{\MLE}{\mathrm{MLE}}

\newcommand{\angleInterval}{\left[-\pi, \pi \right]}
\newcommand{\cdist}[1]{\mathcal{#1}}
\newcommand{\cdGaussian}{\cdist{N}}
\newcommand{\cdMises}{\cdist{M}}

\newcommand{\func}[2]{#1\!\mtup{#2}}

\NewDocumentEnvironment{toggleitemize}{s}
{
  \IfBooleanTF {#1} {%
    \renewcommand{\item}{}%
    \renewenvironment{itemize}{}{}%
  }{
    \begin{itemize}
    }
  }
  {
    \IfBooleanTF {#1} {%
    }{
    \end{itemize}
  }
}

\newcommand{\obj}{O}
\newcommand{\pose}{\mvec{P}}
\newcommand{\rel}{R}
\newcommand{\srel}{s}
\newcommand{\srelset}{\styleset{S}}
\newcommand{\reltrg}{u}
\newcommand{\relref}{v}
\newcommand{\relrefs}{\styleset{V}}

\newcommand{\pos}{\mvec{p}}

\newcommand{\orimat}{\mvec{Q}}

\newcommand{\objset}{\styleset{\obj}}
\newcommand{\poseset}{\styleset{\pose}}

\newcommand{\objname}{\eta}

\newcommand{\objgeometry}{g}

\newcommand{\baretimekx}[1][k]{{t_{#1}}}
\newcommand{\baretimek}{{\baretimekx}}
\newcommand{\timekx}[1][k]{{\left( \baretimekx[#1] \right)}}
\newcommand{\timek}{{\timekx[k]}}

\newcommand{\trg}{{\timekx[k+1]}}

\newcommand{\trgrelation}{\rel^*}
\newcommand{\trgsrel}{{\srel^*}}
\newcommand{\trgobj}{{\reltrg^*}}
\newcommand{\trgrefs}{{\relrefs^*}}
\newcommand{\trgpos}{\pos^\trg_\trgobj}
\newcommand{\trgpose}{\pose^\trg_\trgobj}

\newcommand{\relationfull}{\rel = \mtup{ \srel, \reltrg, \relrefs }}
\newcommand{\trgrelationfull}{\trgrelation = \mtup{ \trgsrel, \trgobj, \trgrefs }}

\newcommand{\genmodel}{G}

\newcommand{\setOneToN}[1][n]{\mset{1, \dots, #1}}

\newcommand{\command}{C}

\newcommand{\cRadius}{r}
\newcommand{\cAzim}{\angl}
\newcommand{\cHeight}{h}

\newcommand{\cdParam}{\theta}

\newcommand{\cdParamAngle}{\cdParam_\angl}

\newcommand{\cdParamRH}{\cdParam_{\cRadius \cHeight}}

\newcommand{\cDistrib}{\mathcal{C}}

\newcommand{\cMeanRH}{\mvec{\mu}_{\cRadius \cHeight}}
\newcommand{\cCovRH}{\mvec{\Sigma}_{\cRadius \cHeight}}

\newcommand{\cylpos}{\mvec{c}}

\newcommand{\sample}{D}
\newcommand{\genmodelset}{{\mathcal{G}}}
\newcommand{\samplesetall}{{\mathcal{D}}}
\newcommand{\samplesetrel}{\mathcal{D}_\srel}
\newcommand{\srelj}{{\srel j}}

\newcommand{\baresampletimeini}{\baretimekx[{k_\srelj}]}
\newcommand{\baresampletimetrg}{\baretimekx[{k_\srelj+1}]}
\newcommand{\sampletimeini}{{\timekx[{k_\srelj}]}}
\newcommand{\sampletimetrg}{{\timekx[{k_\srelj+1}]}}

\newcommand{\nobjk}{n^\timek}
\newcommand{\numknownobjects}{N}

\begin{document}
\nolinenumbers
  
\onecolumn
\firstpage{1}

\title[Interactive and Incremental Learning of Spatial Object Relations]{Interactive and Incremental Learning of Spatial Object Relations from Human Demonstrations}

\author[\firstAuthorLast ]{\Authors} 
\address{} 
\correspondance{} 

\extraAuth{}

\maketitle

\begin{abstract}

Humans use semantic concepts such as spatial relations between objects 
to describe scenes and communicate tasks such as
``Put the tea to the right of the cup''
or ``Move the plate between the fork and the spoon.''
Just as children, assistive robots must be able to learn the sub-symbolic meaning of such concepts from human demonstrations and instructions.
We address the problem of 
incrementally learning geometric models of spatial relations 
from few demonstrations 
collected online during interaction with a human.
Such models enable a robot to manipulate objects
in order to fulfill desired spatial relations specified by verbal instructions.
At the start, we assume the robot has no geometric model of spatial relations.
Given a task as above, the robot requests the user to demonstrate the task once in order to create a model from a single demonstration,
leveraging cylindrical probability distribution as generative representation of spatial relations.
We show how this model can be updated incrementally with each new demonstration without access to past examples 
in a sample-efficient way
using incremental maximum likelihood estimation,
and demonstrate the approach on a real humanoid robot.

\tiny
\keyFont{
  \section{Keywords:} Cognitive Robotics, Learning Spatial Object Relations, Semantic Scene Manipulation, Incremental Learning, Interactive Learning
}
\end{abstract}

\section{Introduction}

While growing up, humans show impressive capabilities to continually learn intuitive models of the physical world 
as well as concepts which are essential to
communicate and interact with others.
While an understanding of the physical world can be created through exploration,  concepts such as the meaning of words and gestures are learned by observing and imitating others.
If necessary, humans give each other explicit explanations and demonstrations to purposefully help the learner improve their understanding of a specific concept.
These can be requested by the learner after acknowledging their incomplete understanding, 
or by the teacher when observing a behavior that does not match their internal model~\citep{Grusec1994SocialLearningTheory}.
Assistive robots that naturally interact with humans and support them in their daily lives 
should be equipped with such continual and interactive learning abilities, allowing them to improve their current models and learn new concepts from their users interactively and incrementally.

One important class of concepts children need to learn are the meanings of spatial prepositions such as \emph{right of}, \emph{above} or \emph{close to}. 
Such prepositions define geometrical relationships between spatial entities~\citep{OKeefe2003VectorGrammarPlaces}, such as 
objects, living beings or conceptual areas,
which are referred to as \emph{spatial relations}~\citep{Stopp1994UtilizingSpatialRelations,Aksoy2011LearningSemanticsObject,Rosman2011LearningSpatialRelationships}.
Spatial relations play an important role in communicating manipulation tasks in natural language, \eg, in 
``Set the table by placing a plate \emph{on} the table, the fork to \emph{the left} of the plate, and the knife to the \emph{right of} the plate.''
By abstracting from precise metric coordinates and the involved entities' shapes, spatial relations allow the expression of tasks on a semantic, symbolic level.
However, a robot performing such a task must be able to derive 
subsymbolic placing positions that are needed to parameterize actions.
Such \emph{signal-to-symbol gap} remains a grand challenge in cognitive robotics~\citep{Kruger2011ObjectActionComplexes}.
Just like a child, a robot should be able to learn such mapping of spatial object relations 
from demonstrations provided by humans.

\begin{figure}[tb]
  \centering
  \includegraphics[width=\linewidth]{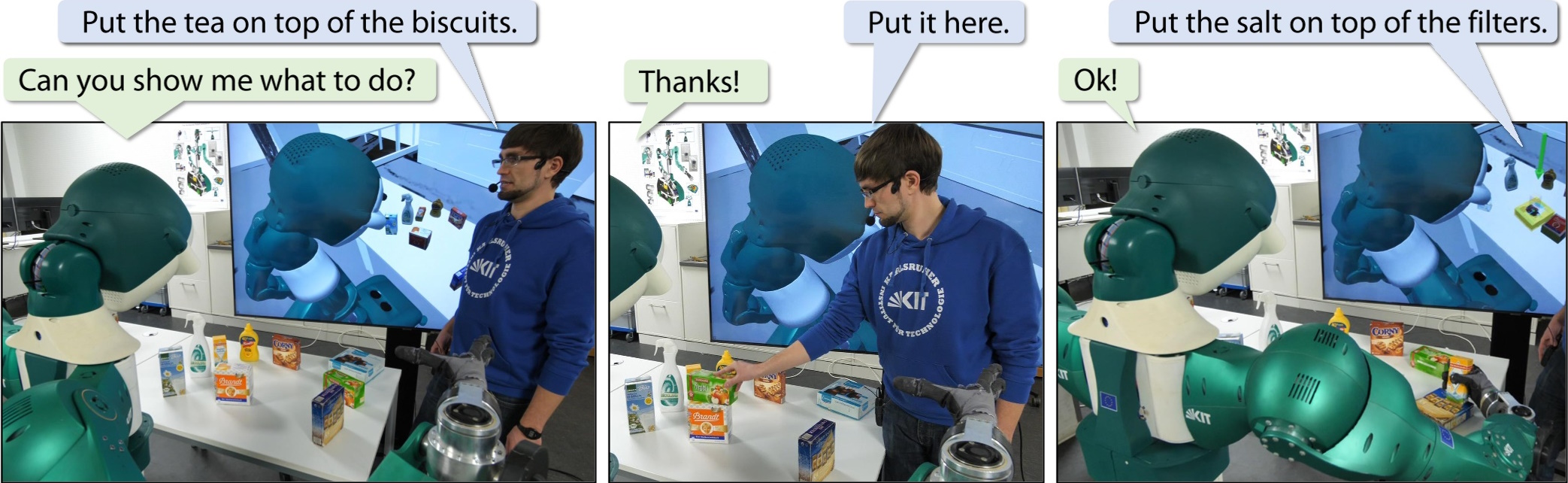}
  \caption{
    Incremental learning of spatial relations
    from human demonstrations.
  }
  \label{fig:face}
\end{figure}

In this work, we consider a robot that has no prior knowledge about the geometric meaning of any spatial relations yet. 
When given the task to manipulate a scene to fulfill a desired spatial relation between two or more objects, such as a cup \emph{in front of} a bottle,
the robot should request a demonstration from the user if it has no model of the spatial relation or if its current model is insufficient (\cref{fig:face}). Similarly, the robot should be able to receive corrections from the human after executing the task. 
Finally, having received a new demonstration, the robot should 
be able to derive a model of the spatial relation from the very first sample and subsequently update its model incrementally with each new demonstration, \ie, without the need to retrain the model with all previously observed demonstrations \citep{Losing2018IncrementalOnLineLearning}.

These goals pose hard requirements for the underlying representation of spatial relations and the cognitive system as a whole.
The robot needs to inform the user in case it cannot perform the task by asking for help while maintaining an internal state of the interaction.
In addition, the robot will only receive very sparse demonstrations -- every single demonstration should be used to update the robot's model of the spatial relation at hand.
As a consequence, we require a very sample-efficient representation that can be constructed from few demonstrations and incrementally updated with new ones.

Obtaining a sample-efficient representation can be achieved by introducing
bias about the hypothesis space
\citep{Mitchell1982GeneralizationSearcha}.
Bias reduces the model's capacity, and thereby its potential variance, 
but more importantly, also reduces the amount of data required to train the model.
This effect is also known as the bias-variance tradeoff.
Compared to a partially model-driven approach with a stronger bias, a purely data-driven black-box model can offer superfluous capacity, slowing down training.
Bias can be introduced
by choosing a model whose structure matches that of the problem at hand.
For the problem of placing objects according to desired spatial relations,
we proposed to represent spatial relations
as parametric probability distributions defined in cylindrical coordinates in our previous work 
\citep{Kartmann2020RepresentingSpatialObject,Kartmann2021SemanticSceneManipulation},
observing that capturing relative positions in terms of horizontal distance, horizontal direction, and vertical distance closely matches the notions of common spatial prepositions.

An interesting consideration arises when learning a model from a single demonstration.
A single example does not hold any variance; 
consequently, the learner's only option is to reproduce this demonstration as closely as possible.
When receiving more examples, the learner can add variance to its model,
thus increasing its ability to adapt to more difficult scenarios.
This principle is leveraged by version space algorithms
\citep{Mitchell1982GeneralizationSearcha}, 
which have been used in robotics to incrementally learn task precedence \citep{Pardowitz2005LearningSequentialConstraints}.
While our approach is not 
a version space algorithm
per se, it behaves similarly with respect to incremental learning: 
The model starts with no variance and acquires more variance with new 
demonstrations,
which increases its ability to handle more difficult scenarios
such as cluttered scenes.

We summarize our contributions as follows:
\begin{enumerate}
  \item 
  We present an approach for a robot interacting with a human
  that allows the robot to
  (a)~request demonstrations from a human 
  for how to manipulate the scene according to desired spatial relations specified by a language instruction
  if the robot has not sufficient knowledge about how to perform such a task,¸
  as well as
  (b)~to use corrections after each execution 
  to continually improve its internal model about spatial relations.
  
  \item
  We show how our representation of spatial relations based on cylindrical distributions proposed in
  \citep{Kartmann2021SemanticSceneManipulation} 
  can be incrementally learned from few demonstrations 
  based only on the current model and a new demonstration
  using incremental maximum likelihood estimation.
  
\end{enumerate}

We evaluate our approach in simulation and real world experiments on the humanoid robot \armarVI~\citep{Asfour2019}\footnote{\url{https://youtu.be/x6XKUZd_SeE}}.

\section{Related Work}
\label{s:related-work}

\newcommand{\citev}[1]{\textcolor{gray}{(#1)}~\citep{#1}}

In this section, we discuss related work in the areas of using spatial relations in the context of human-robot interaction 
and 
the incremental learning of spatial relation models.

\subsection{Spatial Relations in Human-Robot Interaction and Learning from Demonstrations}

\begin{toggleitemize}*
  \item Spatial relations have been used to enrich language-based human-robot interaction.
  \item Many works focus on using spatial relations to resolve referring expressions identifying objects \citep{Bao2016TeachRobotsUnderstanding,Tan2014GroundingSpatialRelations}
  as well as locations \citep{Fasola2013UsingSemanticFields,Tellex2011UnderstandingNaturalLanguage}
  for manipulation and navigation.
  \item In these works, the robot passively tries to parse the given command or description without querying the user for more information if necessary.
  As a consequence, if the sentence is not understood correctly, the robot is unable to perform the task.
  \item Other works use spatial relations in dialog to resolve such ambiguities.
  \item \cite{Hatori2018InteractivelyPickingRealWorld}
  query for additional expressions if the resolution did not score a single object higher by a margin than all other objects.
  \item \cite{Shridhar2018InteractiveVisualGrounding,Shridhar2020INGRESSInteractiveVisual} and \cite{Dogan2022AskingFollowUpClarifications} 
  formulate clarification questions describing candidate objects using, among others, spatial relations between them.
  \item 
  These works use spatial relations in dialog to identify objects which the robot should interact with. 
  However, our goal is to perform a manipulation task defined by desired spatial relations between objects. 
\end{toggleitemize}

\begin{toggleitemize}*
  \item A special form of human-robot interaction arises in the context of robot learning from human demonstrations
  \citep{Ravichandar2020RecentAdvancesRobot}.
  \item There, the goal is to teach the robot a new skill or task instead of performing a command.
  \item Spatial relations have been used in such settings to specify parameters of taught actions.
  \item Similar to the works above, given the language command and current context, \cite{Forbes2015RobotProgrammingDemonstration} resolve referring expressions to identify objects using language generation to find the most suitable parameters for a set of primitive actions.
  \item Prepositions from natural language commands are incorporated as action parameters in a task representation based on part-of-speech tagging by \cite{Nicolescu2019LearningComplexStructuredTasks}.
  \item 
  \item 
    These works focus on learning the structure of a task including multiple actions and their parameters.
    However,
    action parameters are limited to a finite set of values, and spatial relations are implemented as fixed position offsets. 
    In contrast, our goal is learning continuous, geometric models of the spatial relations themselves.
  
\end{toggleitemize}

\subsection{Learning Spatial Relation Models}

\begin{toggleitemize}*
  \item 
  Many works have introduced models to classify existing spatial relations between objects 
  to improve scene understanding 
  \citep{%
    Rosman2011LearningSpatialRelationships%
    ,Sjoo2011LearningSpatialRelations%
    ,Fichtl2014LearningSpatialRelationships%
    ,Yan2020RoboticUnderstandingSpatial%
  } 
  and human activity recognition 
  \citep{%
    Lee2020QSRNetEstimatingQualitative%
    ,Zampogiannis2015LearningSpatialSemantics%
    ,Dreher2020LearningObjectActionRelations%
  }.
  \item 
  These models are either hand-crafted or are not learned incrementally.
  \item In contrast, our models are learned incrementally from demonstrations collected during interaction.
  \item 
  Few works consider models for classifying spatial relations which could be incrementally updated with new data.
  \item \cite{Mees2020LearningObjectPlacements} train a neural network model to predict a pixel-wise probability map of placement positions given a camera image of the scene and an object to be placed according to a spatial relation.
  However, their models are not trained incrementally, and training neural networks incrementally is, in general, not trivial \citep{Losing2018IncrementalOnLineLearning}. 
  \item In an earlier work, \cite{Mees2017MetricLearningGeneralizing} propose a metric learning approach to model spatial relations between two objects represented by point clouds. 
  The authors learn a distance metric measuring how different the realized spatial relations in two scenes are.
  Recognizing the spatial relation in a given scene is then reduced to a search of known examples that are similar to the given scene according to the learned metric. 
  Once the metric is learned and kept fixed, this approach inherently allows adding new samples to the knowledge base, which potentially changes the classification of new, similar scenes. 
  However, their method requires storing all encountered samples to keep a notion of known spatial relations,
  while our models can be updated incrementally
  with a limited budget of stored examples \citep{Losing2018IncrementalOnLineLearning}.
  \item \cite{Mota2018IncrementallyGroundingExpressions} follow a related idea to learn classification models of spatial relations incrementally.
  They encode spatial relations as 1D and 2D histograms over the relative distances or directions (encoded as azimuth and elevation angles), of points in two point clouds representing two objects. 
  These histograms can be incrementally updated by merely adding a new observation to the current frequency bins.
   
  \item 
  However, all of these models are \emph{discriminative}, 
  \ie, they determine the existing relations between two objects in the current scene.
  \item
  In contrast, 
  our goal is to \emph{generate} a new target scene
  given desired spatial relations. 
  While discriminative models can still be applied
  by exhaustively sampling the solution space 
  (\eg, possible locations of manipulated objects),
  classifying the relations in these candidates
  and choosing one that contains the desired relation,
  we believe
  that it is more effective to directly learn and apply
  generative geometric models of spatial relations.
  \item 
  In our previous works, we introduced generative representations of 
  3D spatial relations
  in the form of parametric probability distributions over placing positions~\citep{Kartmann2021SemanticSceneManipulation}.
  \item These probabilistic models 
  can be sampled to obtain suitable placing positions 
  for an object to fulfill a desired spatial relation 
  to one or multiple reference objects.
  \item We have shown how these models can be learned from human demonstrations which were collected offline. 
  \item In this work, we  show how demonstrations can be given interactively and how the models can be updated in a fully incremental manner, \ie, relying solely on the current model and a new demonstration.
\end{toggleitemize}

\section{Problem Formulation and Concept}
\label{s:problem-formulation}

In the following, we formulate the problem of interactive and incremental learning of spatial relation models from human demonstrations 
and introduce the general concept of the work. 
  In \cref{ss:problem:scene-manipulation}, we summarize the actual task of semantic scene manipulation which the robot has to solve.
  In \cref{ss:problem:semantic-memory}, we describe the semantic memory system as part of the entire cognitive control architecture used on the robot.
  In \cref{ss:problem:incremental-learning}, we formulate the problem of incremental learning of spatial relation models.
  In \cref{ss:problem:interactive-teaching}, we describe the envisioned human-robot interaction task and explain how we approached each subproblem in \cref{s:approach}.

\subsection{Semantic Scene Manipulation}
\label{ss:problem:scene-manipulation}

We consider the following problem:
Given a scene with a set of objects and a language command specifying  spatial relations between these objects,
the robot must transfer the initial scene to a new scene fulfilling the specified relations by executing an action of a set of actions.
We denote points in time as 
$ \baretimekx[0], \dots, \baretimekx[k], \ldots
  \ \left(\baretimekx[k] \in \reals\right)
$
that can be viewed as events where the robot is given a command or makes an observation.
The scene model is part of the robot's working memory and contains 
the current configuration 
\begin{equation}
\poseset^\timek = \mset{ \pose_1^\timek, \dots, \pose_n^\timek }   
\end{equation}
of $\nobjk$ objects at time $\baretimek$, 
where \mbox{$\pose_i^\timek \in \SEthree$} is the pose of object $i$ with position~$\pos_i^\timek \in \reals^3$ and orientation~$\orimat_i^\timek \in \SOthree$. 
$\SEthree$ and $\SOthree$ denote the special Euclidean group and special orthogonal group, respectively.
The desired relation 
\begin{equation}
  \label{eq:grounding}
  \trgrelationfull
  \leftarrow
  \func{\mathrm{ground}}{ \command }
\end{equation}
is obtained by parsing and grounding the natural language command $\command$,
\ie, extracting the phrases referring to objects and relations and mapping them to the respective entities in the robot's working and long-term memories%
\footnote{Please refer to \cite{Kartmann2021SemanticSceneManipulation} for more details on the  natural language understanding part.}.
This desired relation consists of 
a symbol 
$\trgsrel \in \srelset$ 
describing the identity of the relation,
a \emph{target} object~$\trgobj \in \setOneToN[\nobjk]$ 
and a set of \emph{reference} objects~%
$ \trgrefs \subseteq \setOneToN[\nobjk] \setminus \mset{ \trgobj }$.
$\srelset$ is the set of known spatial relation symbols. 
The robot's task is to place the target object $\trgobj$ at an appropriate pose
$\trgpose$ which fulfills the desired spatial relation $\trgrelation$.
We aim at keeping the object's original orientation,
therefore this task is reduced to finding a suitable position
$\trgpos$.

Our approach to finding suitable placing positions is based on a generative model~$\genmodel$ of spatial relations.
This model is able to generate suitable target object positions fulfilling a relation $R=\mtup{\srel, \reltrg, \relrefs}$ based on the current scene $\poseset^\timek$ 
and semantic object information $\objset$
(object names and geometry, see \cref{eq:object-info} below),
that is formally
\begin{equation}
  \label{eq:generative-model}
  \pos_\reltrg^\trg \sim \func{\genmodel_\srel}{ \reltrg, \relrefs, \objset, \poseset^\timek } 
  .
\end{equation}
The generative models $\genmodel_\srel$ of spatial relations $\srel$ can be learned from human demonstrations. 
In our previous work, we recorded human demonstrations for each spatial relation using real objects and learned the generative models~$\genmodel_\srel$ offline.
Each demonstration consisted of 
the initial scene~$\poseset^\timek$, 
a desired relation \mbox{$\trgrelationfull$} verbalized as a language command for the human demonstrator,
and the resulting scene~$\poseset^\trg$ created by the demonstrator by manipulating the initial scene.
Therefore, each demonstration has the form
\begin{equation}
  \label{eq:sample}
  \sample =
  \mtup{ \poseset^\timek, \rel, \poseset^\trg },  \quad \relationfull
\end{equation}
and can be used to learn the generative model $\genmodel_\srel$ of the relation $\srel$.
In contrast to the previous work, in this work
we consider the problem of 
interactively collecting samples by querying demonstrations from the user 
and incrementally updating the generative models of the spatial relations in the robot's memory with each newly collected sample.

\subsection{Robot Semantic Memory}
\label{ss:problem:semantic-memory}

The robot's semantic memory consists of two parts: 
the \emph{prior knowledge}, \ie, information defined a-priori by a developer,
and the \emph{long-term memory}, \ie, experience gathered by the robot itself \citep{Asfour2017KarlsruheARMARHumanoid}.
In our scenario,
the prior knowledge contains
semantic information about $\numknownobjects$ known objects,
\begin{equation}
  \label{eq:object-info}
  \objset = \mset{\obj_i}_{i=1}^\numknownobjects = \mset{\mtup{ \objgeometry_i, \objname_i }}_{i=1}^\numknownobjects ,
\end{equation}
including object names $\objname_i$ and 3D models $\objgeometry_i$,
as well as
names of spatial relations,
so that language phrases referring to both objects and relations can be grounded in entities stored in the robot's memory by natural language understanding
as indicated in~\cref{eq:grounding}. 
We assume no prior knowledge about the geometric meaning of the relations.
Instead, this geometric meaning is learned by the robot during interaction in a continual manner,
and is thus part of the long-term memory.
In other words, the long-term memory contains generative models $\genmodel_\srel^{\timek}$ of spatial relations $\srel \in \srelset^\timek$
representing the robot's current understanding of spatial relations at time $\baretimek$,
\begin{equation}
  \genmodelset^\timek = \mset{\left. \genmodel_\srel^\timek \right|  \srel \in \srelset^{\timek}},
\end{equation}
where
$\srelset^\timek \subseteq \srelset$~is the set of spatial relations for which the robot has learned a generative model at~$\baretimek$.
These models are based on samples collected from human demonstrations $\samplesetall$, which are contained in the long-term memory as well:
\begin{equation}
  \label{eq:sample-set-all}
  \begin{aligned}
    \samplesetall^\timek &= \mset{\samplesetrel^\timek \left| \ \srel \in \srelset^\timek \right.},
    \\ 
    \samplesetrel^\timek &= 
    \mset{ \sample_\srelj }_{j=1}^{m_\srel^\timek} 
    = \mset{ 
      \poseset^\sampletimeini_\srelj, 
      \rel_\srelj, 
      \poseset^\sampletimetrg_\srelj 
    }_{j=1}^{m_\srel^\timek},
  \end{aligned}
\end{equation}
where $\samplesetrel^\timek$ is the set of $m_\srel^\timek$ collected samples of spatial relation~$\srel$ at time $\baretimek$,
and $\baresampletimeini, \baresampletimetrg$ refer to the time a sample was collected.
At the beginning, the robot's long-term memory is empty, therefore
\begin{equation}
  \genmodelset^{\timekx[0]} = \samplesetall^{\timekx[0]} = \srelset^{\timekx[0]} = \emptyset
  .
\end{equation}

\subsection{Learning Spatial Relations: Batch vs Incremental}
\label{ss:problem:incremental-learning}

During interactions, the robot will collect new samples from human demonstrations.
When the robot receives a new demonstration for relation $\srel$ at time~$\baretimekx[k]$ $(k > 0)$, it first stores the new sample $\sample_\srel$ in its long-term memory:
\begin{equation}
  \samplesetrel^{\timek} \leftarrow 
  \mset{\sample_{\srel}} \cup \begin{cases}
    \samplesetrel^{\timekx[k-1]}, & \srel \in \srelset^{\timekx[k-1]}
    \\ 
    \emptyset, & \text{else}
  \end{cases}
\end{equation}
Afterwards, the robot may query its long-term memory for the relevant samples $\samplesetrel^{\timek}$ collected to date 
and use them to update its model $\genmodel_\srel$ of $\srel$:
\begin{equation}
  \label{eq:update-model}
  \genmodel_\srel^\timek \leftarrow
  \func{\mathrm{update}}{
    \genmodel_\srel^{\timekx[k-1]}, 
    \samplesetrel^{\timek}
  }
\end{equation}
Note that such an incremental model update with each new sample requires a very sample-efficient representation.
We refer to \cref{eq:update-model} as the \emph{batch update problem}.
The model is expected to adapt meaningfully to each new sample, but all previous samples are needed for the model update.
In the machine learning community,
incremental learning can be defined as proposed by
\cite{Losing2018IncrementalOnLineLearning}:
\begin{quote}
  ``We define an incremental learning algorithm as one that generates on a given stream of training data $s_1, s_2, \dots, s_t$ a sequence of models $h_1, h_2, \dots, h_t$. 
  In our case [...] $h_i : \reals^n \rightarrow \{1, \dots, C\}$ is a model function solely depending on $h_{i-1}$ and the recent $p$ examples $s_i, \dots, s_{i-p}$, with $p$ being strictly limited.'' 
\end{quote}
Comparing this definition to \cref{eq:update-model}, it becomes clear that it is not an instance of incremental learning in this sense, as the number of samples $|\samplesetrel|$ in the memory is not bounded by a constant. 
This raises the question of whether representations of spatial relations in the form of cylindrical distributions as proposed in our previous work can be learned in a truly incremental way using a limited budget of stored samples.
In this work, we further investigate the question of incremental learning of spatial relations without even storing \emph{any} samples, \ie, solely based on the current model $\genmodel_\srel^{\timekx[k-1]}$ and the latest sample~$\sample_{\srel}$, forming the \emph{incremental update problem}:
\begin{equation}
  \label{eq:update-model-fully-incremental}
  \genmodel_\srel^\timek \leftarrow
  \func{\mathrm{update}}{
    \genmodel_\srel^{\timekx[k-1]}, 
    \sample_{\srel}
  }
\end{equation}

\subsection{Interactive Learning of Spatial Relations}
\label{ss:problem:interactive-teaching}

We now describe a scenario of a robot interacting with a human
where the human
gives a semantic manipulation command to the robot
while the robot can gather new samples from human demonstrations.
The scheme is illustrated in~\cref{fig:interaction}.
The procedure can be described as follows:

\begin{figure}[tb]
	\centering
  \includegraphics[width=0.5\linewidth]{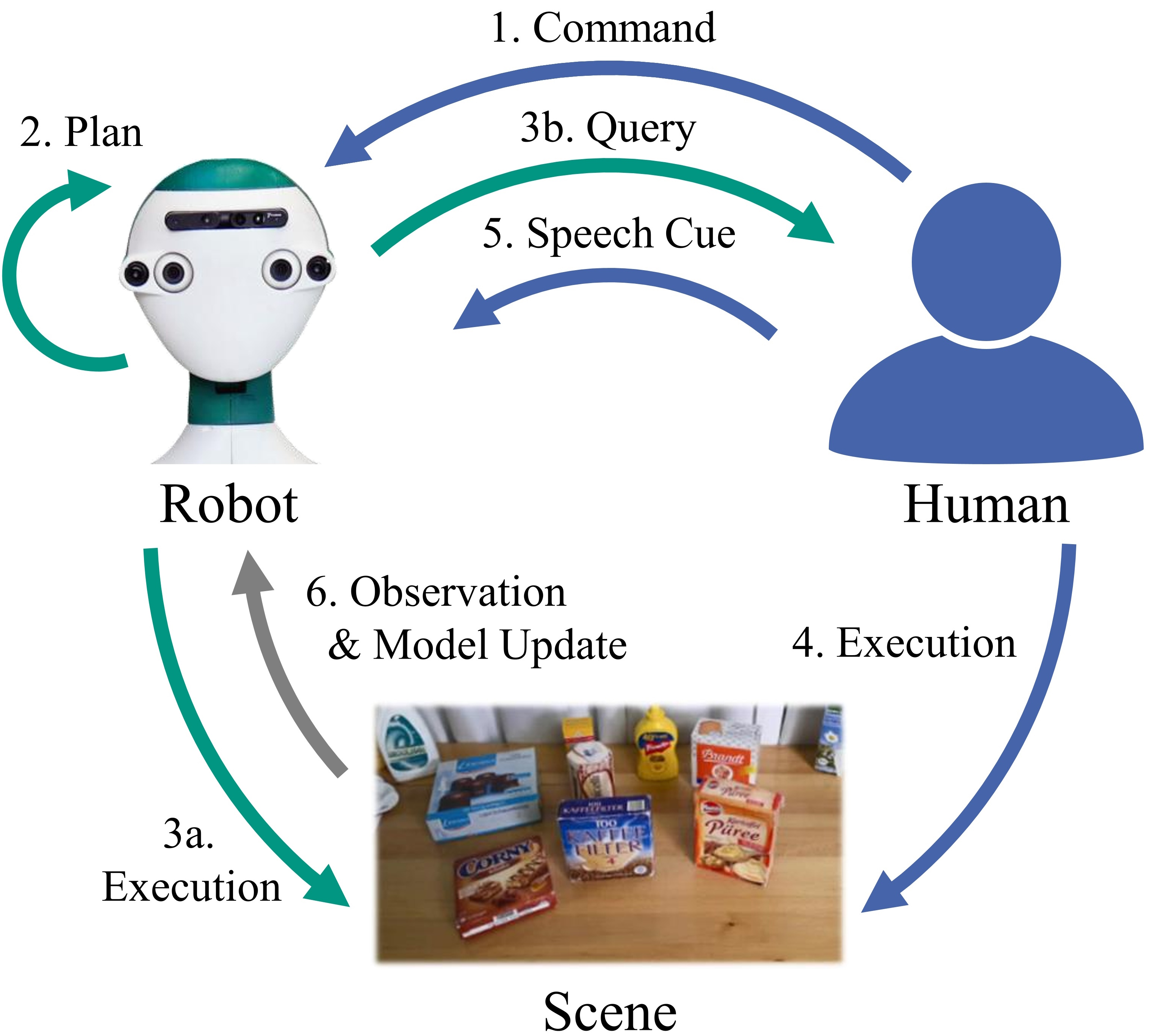}
	\caption{
		Scheme for a robot interacting with a human
    to learn geometric models of spatial relations.
	}
	\label{fig:interaction}
\end{figure}

\begin{enumerate}
  \item 
  The human gives a command to the robot at~$\baretimek$, specifying a spatial relation $\trgrelationfull$ as in~\cref{eq:grounding}.
  
  \item 
  The robot observes the current scene $\poseset^\timek$ and plans the execution of the given task using its current model~$\genmodel_{\trgsrel}^{\timekx[k]}$.
  Planning may be successful or fail due to different reasons. 
  Assuming the task given by the user is solvable, the failure can be attributed to an insufficient model.
  
  \item 
  Depending on the outcome of 2.:
  \begin{enumerate}
    \item[3a.] 
    \emph{Planning is successful:}
    The robot found a suitable placing position and executes the task by manipulating the scene.
    If the execution was successful and the user is satisfied, the interaction is finished.
    \item[3b.] 
    \emph{Planning fails:}
    The robot's current model is insufficient, 
    thus, it queries the human for a demonstration of the task.
  \end{enumerate}
  
  \item
  The human was queried for a demonstration~(3b.) 
  or wants to correct the robot's execution~(3a.).
  In both cases, the human performs the task by manipulating the scene.
  
  \item 
  The human signals that the demonstration is complete by giving a speech cue (\eg, ``Put it here'') to the robot.
  
  \item 
  When receiving the speech cue, the robot observes the changed scene $\poseset^\trg$,
  creates a new sample 
  $ \sample =
  \mtup{ \poseset^\timek, \trgrelation, \poseset^\trg }$,
  stores the sample in the long-term memory
  and updates its model to obtain the new model~$\genmodel_{\trgsrel}^{\timekx[k+1]}$ 
  as described in~\cref{ss:problem:incremental-learning}.
  
\end{enumerate}

\section{Methods and Implementation}
\label{s:approach}

To solve the underlying task of semantic scene manipulation, we rely on our previous work 
which is briefly described in \cref{ss:approach:cylindrical-distribution}.
We outline the implementation of the robot's semantic memory in \cref{ss:approach:memory}.
We describe how each new sample is used to update a spatial relation's model in \cref{ss:approach:incremental-learning}.
Finally, we explain how we implement the defined interaction scenario in \cref{ss:approach:interactive-teaching}.

\subsection{3D Spatial Relations as Cylindrical Distributions}
\label{ss:approach:cylindrical-distribution}

In \citep{Kartmann2021SemanticSceneManipulation}, we proposed a model of spatial relations based on a cylindrical distribution~$\cDistrib$ 
\begin{equation}
  \label{eq:cylindrical-def}
  \mtup{\cRadius, \cAzim, \cHeight} \sim \func{\cDistrib}{\cdParam},
  \quad \cdParam = \mtup{\cdParamRH, \cdParamAngle}
\end{equation}
over the cylindrical coordinates 
radius $\cRadius \in \reals_{\geq 0}$,
azimuth $\cAzim \in \angleInterval$
and height $\cHeight \in \reals$.
The radius and height follow a joint Gaussian distribution $\cdGaussian$,
while the azimuth, as an angle, follows
a von Mises distribution $\cdMises${} (which behaves similar to a Gaussian distribution but is defined on the unit circle),
\begin{equation}
  \label{eq:radius-height-gaussian}
  \mtup{\cRadius, \cHeight} 
  \sim \func{\cdGaussian}{\cdParamRH}, 
  \quad \cdParamRH = \mtup{ \cMeanRH, \cCovRH }, 
  \qquad
  \angl \sim \func{\cdMises}{\cdParamAngle}, 
  \quad \cdParamAngle = \mtup{ \mu_\angl, \kappa_\angl }
\end{equation}
so the joint probability density function of $\cDistrib$ is given by
\begin{equation}
  \label{eq:cylindrical-pdf}
  \func{p_\cdParam}{\cRadius, \cAzim, \cHeight}
  = \func{p_{\cdParamRH}}{\cRadius, \cHeight} \cdot \func{p_{\cdParamAngle}}{\cAzim}
\end{equation}
with $\func{p_{\cdParamAngle}}{\angl}$ and $\func{p_{\cdParamRH}}{\cRadius, \cHeight}$
being the respective probability density functions of the distributions in
\cref{eq:radius-height-gaussian}.
We claim that cylindrical coordinates are a suitable space for representing spatial relations as they inherently encode horizontal distance (\emph{close to}, \emph{far from}), 
direction (\emph{left of}, \emph{behind}) 
and vertical distance (\emph{above}, \emph{below}),
which are the qualities many spatial relations used by humans are defined~by.
  Therefore, we leverage a cylindrical distribution as a distribution over suitable placing positions of a target object relative to one or more reference objects.
  The corresponding cylindrical coordinate system is centered at the bottom-projected centroid of the axis-aligned bounding box enclosing all reference objects
  and scales with the size of that bounding box.
  By defining the cylindrical coordinate system based on the reference objects' joint bounding box, 
  we can apply the same spatial relation models to single and multiple reference objects.
  Especially, this allows considering spatial relations that inherently involve multiple reference objects (\eg, \emph{between}, \emph{among}).

\subsection{Initialization of Robot's Semantic Memory}
\label{ss:approach:memory}

We build our implementation of the cognitive architecture of our robot in \armarx \citep{Vahrenkamp2015RobotSoftwareFramework,Asfour2017KarlsruheARMARHumanoid}. The architecture consists of three-layer for 
1) low-level sensorimotor control, 
2) high-level for semantic reasoning and task planning and 
3) mid-level as a memory system and mediator between the symbolic high-level and subsymbolic low-level. 
The memory system contains segments for different modalities, such as object instances or text recognized from speech.
A segment contains any number of entities that can receive new observations and thus evolve over time.
Here, we define three new segments:
\begin{itemize}
  \item 
The \emph{sample segment} implements the set of human demonstrations~$\samplesetall$ in \cref{eq:sample-set-all}. 
It contains one entity for each $\srel \in \srelset^\timek$, each one representing ``all collected samples of relation~$\srel$.''
New samples are added as new observations to the respective entity.
Thus, $\samplesetrel^\timek$ in \cref{eq:sample-set-all} is obtained by querying the sample segment for all observations of entity $\srel$.
  
  \item 
The \emph{spatial relation segment} contains the robot's knowledge about spatial relations.
There is one entity per $\srel \in \srelset$, which holds the information from prior knowledge such as the relations' names for language grounding and verbalization.
In addition, each entity contains the current geometric model~$\genmodel_\srel^\timek = \cdParam $ with $\cdParam$~as in \cref{eq:cylindrical-def}.
  
  \item 
The \emph{relational command} segment contains the semantic manipulation tasks $\trgrelationfull$ extracted from language commands~\cref{eq:grounding}.
The latest observation is the current or previous task.

\end{itemize}

In the beginning, the sample segment and the relational command segment are initialized to be empty, \ie, there are no collected samples yet and no command has been given yet.
The spatial relation segment is partially initialized from prior knowledge.
However, in accordance with the sample segment, the entities contain no geometric model $\genmodel_\srel$ yet,
\ie~$\genmodel_\srel^{\timekx[0]} = \emptyset$.

\subsection{Incremental Learning of Spatial Relations}
\label{ss:approach:incremental-learning}

\newcommand{\mysubsubsection}[1]{\vspace{.5em} \noindent \textbf{#1} \ }

  Now, we describe how the batch update problem can be solved using cylindrical distributions.
  Then, we show how the same mathematical operations can be performed incrementally, \ie, without accessing past samples.

\mysubsubsection{Batch Updates}
Due to the simplicity of our representation of spatial relations, updating the geometric model of a spatial relation, \ie, its cylindrical distribution, is relatively straightforward.
To implement the batch update \cref{eq:update-model}, we query all samples of the relation of interest collected so far,
perform Maximum Likelihood Estimation (MLE) to obtain the cylindrical distribution's parameters,
\begin{equation}
  \label{eq:update-model-impl}
  \genmodel_\srel^\timek = \cdParam_\srel^\timek \leftarrow
  \func{\mathrm{MLE}}{
    \samplesetrel^{\timek}
  },
\end{equation}
and update the spatial relation segment.

A cylindrical distribution is a combination of a bivariate Gaussian distribution over $(\cRadius, \cHeight)$ and a von Mises distribution over $\cAzim$, see
\cref{eq:radius-height-gaussian}.
Hence, to perform the MLE of a cylindrical distribution in \cref{eq:update-model-impl},
two independent MLEs are performed for the Gaussian distribution $\func{\cdGaussian}{\cdParamRH}$ 
and the von Mises distribution $\func{\cdMises}{\cdParamAngle}$, respectively,
\begin{equation}
  \label{eq:mle-gaussian}
  \cdParamRH^{*} \leftarrow 
  \func{\MLE_\cdGaussian}{\mset{\cRadius_j, \cHeight_j}_{j=1}^{m_\srel^\timek}} ,
  \qquad
  \cdParamAngle^{*} \leftarrow 
  \func{\MLE_\cdMises}{\mset{\cAzim_j}_{j=1}^{m_\srel^\timek}} ,
\end{equation}
where $\mtup{\cRadius_j, \cAzim_j, \cHeight_j}$ 
are the cylindrical of sample $ \sample_j \in \samplesetrel^{\timek}$
(see \citep{Kartmann2021SemanticSceneManipulation} for more details).

Updating the model with each new sample requires a representation that can be generated from few examples, including just one,
which is the case for cylindrical distributions.
However, special attention has to be paid to the case of encountering the first sample ($|\samplesetrel| = 1$).
As a single sample holds no variance,
an estimated distribution would collapse in a single point.
Note that the model's expected behavior
is well-defined: 
It should reproduce the single sample when generating placing positions, which 
often is a valid candidate. 
Because cylindrical distributions generalize to objects of different sizes, the model can be directly applied to new scenes.
The caveat is that the model is not able to provide alternatives if this placing candidate is, \eg, in collision with other objects. 
If more samples are added ($|\samplesetrel| \geq 2$), the model is generated using standard MLE. 
The resulting distribution will have a variance according to the samples, which allows the robot to generate different candidates for placing~positions.

Technically, in order to allow deriving a generative model from a single demonstration while avoiding special cases in the mathematical formulation, 
we perform a small data augmentation step:
After transforming the sample to its corresponding cylindrical coordinates $ \cylpos = (\cRadius, \cAzim, \cHeight)^\top $,
we create $n$ copies $\cylpos_{1}', \dots, \cylpos_{n}'$ of the sample to which we add small Gaussian noise
\begin{equation}
  \cylpos_{i}'
  \leftarrow \cylpos + 
  \mtup{\varepsilon_\cRadius, \varepsilon_\cAzim, \varepsilon_\cHeight}^\top
  , \quad 
  \varepsilon_\cRadius, \varepsilon_\cAzim, \varepsilon_\cHeight \sim \func{\cdGaussian}{0, \sigma_\varepsilon}
\end{equation}
for $i \in \mset{1, \dots, n}$,
and perform MLE on $\mset{\cylpos, \cylpos_{1}', \dots, \cylpos_{n}'}$.
This keeps the distribution concentrated on the original sample while allowing small variance parameters to keep the mathematical formulation simple and computations numerically stable.
In our experiments, we used~$n=2$ and $\sigma_\varepsilon = 10^{-3}$.
Note that the variance created by this augmentation step is usually negligible compared to the variance created by additional demonstrations.

\mysubsubsection{Incremental Updates}
Now, we present how the model update \cref{eq:update-model-impl}
can be performed in an incremental manner, \ie, with access only to the current model and the new sample.
We follow the method by \cite{Welford1962NoteMethodCalculating} for the incremental
calculation of the parameters of a univariate Gaussian distribution.
By applying this method to the multivarate Gaussian and
using a similar method for von Mises distributions to implement the MLEs~\cref{eq:mle-gaussian}, 
we can incrementally estimate cylindrical distributions.

\cite{Welford1962NoteMethodCalculating} proved that 
the mean $\mu$ and standard deviation $\sigma$ of a one-dimensional series $(x_i),$ $x_i \in \reals,$ 
can be incrementally calculated as follows.
Let $n$ be the current time step, $\mu_1 = x_1$ and $\sigma = \frac{1}{n} \; \tilde{\sigma}$,
\begin{alignat}{3}
  \mu_n &= \left(\frac{n - 1}{n}\right) \cdot \mu_{n-1} + \frac{1}{n} \cdot x_n 
  & \quad (n > 1) \; ,
  \\ 
  \tilde{\sigma}_n 
  &= \tilde{\sigma}_{n-1} 
  + \left(\frac{n - 1}{n}\right) \cdot \left(x_n - \mu_{n-1}\right)^2 
  & \quad (n\geq 1) \; .
\end{alignat}
This method can be extended to a multivariate Gaussian $\func{\cdGaussian}{\mvec{\mu}, \mvec{\Sigma}}$ over a vector-valued series~$\mtup{\mvec{x}_i}$ with $\mvec{x} \in \reals^d$.
With analogous conditions as above,
\begin{alignat}{3}
  \label{eq:mle-mean-incr-multivariate-gaussian}
  \mvec{\mu}_n &= \left(\frac{n - 1}{n}\right) \cdot \mvec{\mu}_{n-1} + \frac{1}{n} \cdot \mvec{x}_n 
  & \quad (n > 1) \; ,
  \\
  \tilde{\mvec{\Sigma}}_n 
  &= \tilde{\mvec{\Sigma}}_{n-1} 
  + \left(\frac{n - 1}{n}\right) \cdot \left(\mvec{x}_n - \mvec{\mu}_{n-1}\right) 
  \left(\mvec{x}_n - \mvec{\mu}_{n-1}\right)^\top 
  & \quad (n\geq 1) \; .
\end{alignat}

A von Mises distribution $\func{\cdMises}{\mu, \kappa}$ over angles $\phi_i$ is estimated as follows \citep{Kasarapu2015MinimumMessageLength}.
Let $\mvec{x}_i$
be the directional vectors in the 2D plane corresponding to the directions~$\phi_i$,
and $\bar{r}$ the normalized length of their sum,
\begin{align}
  \mvec{x}_i \coloneqq& \begin{pmatrix} \cos(\phi_i) \\ \sin(\phi_i) \end{pmatrix} \in \reals^2 \quad (i = 1, \dots, n),
  \qquad 
  \bar{r} \coloneqq \frac{\left\| \sum_{i=1}^n \mvec{x}_i \right\|}{n} \; .
\end{align}
The mean angle $\mu$ is the angle corresponding to the mean direction $\tilde{\mvec{\mu}} \in \reals^2$,
\begin{align}
  \label{eq:mle-vonmises-mean}
  \tilde{\mvec{\mu}} 
  =& \begin{pmatrix} \tilde{\mu}_x \\ \tilde{\mu}_y \end{pmatrix} 
  = \frac{1}{n \cdot \bar{r}} \sum_{i=1}^n \mvec{x}_i,  
  \qquad
  \mu = \func{\tan^{-1}}{\frac{\tilde{\mu}_y}{\tilde{\mu}_x}} .
\end{align}  
The concentration $\kappa$ is computed as the solution of
\begin{align}
  \label{eq:mle-vonmises-kappa}
    \func{A_d}{\kappa} = \bar{r} 
    \; , \qquad
    \text{where} \
    \func{A_d}{\kappa} \coloneqq  \frac{I_{d/2}(\kappa)}{I_{d/2-1}(\kappa)} 
    \; ,
\end{align}
$\func{I_s}{\kappa}$ denotes the modified Bessel function of the first kind and order $s$, 
and
$d$ denotes the dimension of vectors $\mvec{x} \in \reals^d$ on the $(d-1)$-dimensional sphere $\mathbb{S}^{d-1}$
(since we consider the circle embedded in the 2D plane, $d=2$ in our case).
\Cref{eq:mle-vonmises-kappa} is usually solved using closed-form approximations \citep{Sra2012ShortNoteParameter,Kasarapu2015MinimumMessageLength}. 
However, the important insight here is that \cref{eq:mle-vonmises-kappa} does not depend on the values~$\mvec{x}_i$, but only on the normalized length $\bar{r}$ of their \emph{sum}.
This means that the MLE for both $\mu$ and $\kappa$ of a von Mises distribution only depends on the sum
$ \mvec{r} \coloneqq \sum_{i=1}^n \mvec{x}_i $ 
of the directional vectors $\mvec{x}_i$, 
which can easily be computed incrementally.
With $\mvec{r}_1 = \mvec{x}_1$, 
\begin{alignat}{3}
  \mvec{r}_n &
  = \mvec{r}_{n-1} + \mvec{x}_n 
  && \quad (n > 1) , 
  \\
  \qquad
  \bar{r} &
  = \frac{\left\| \mvec{r}_n \right\|}{n}  \; ,
  \quad
  \tilde{\mvec{\mu}} = \frac{1}{n} \; \mvec{r}_n 
  && \quad (n \geq 1),
\end{alignat}
and the remaining terms as in \cref{eq:mle-vonmises-mean,eq:mle-vonmises-kappa}.
Overall, this allows to estimate cylindrical distributions fully incrementally.
Note that the batch and incremental updates are mathematically equivalent and thus yield the same results.

\subsection{Interactive Teaching of Spatial Relations}
\label{ss:approach:interactive-teaching}

\begin{figure}[tb]
  \centering
  \includegraphics[width=0.9\linewidth]{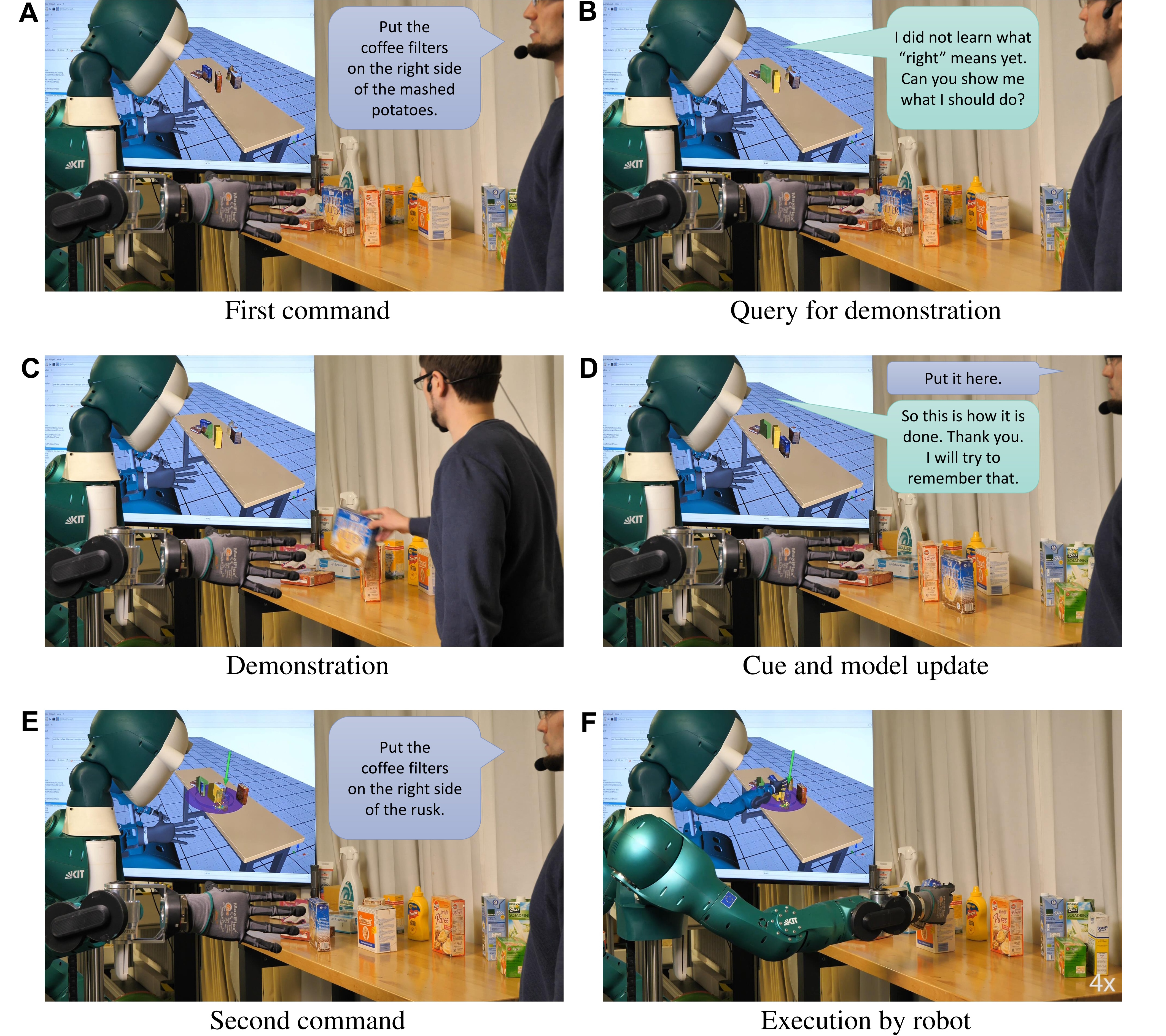}
  \caption{
    Interactively teaching the humanoid robot \armarVI
    to place objects \emph{to the right of} other objects. 
    After the first command by the human (A), 
    the robot queries a demonstration (B). 
    The human demonstrates the task (C) 
    which is used by the robot to create or update its model (D). 
    When given the next command (E), 
    the robot can successfully perfom the task (F).
  }
  \label{fig:experiment-real-robot}
\end{figure}

Finally, we explain how we implement the different steps of the interaction scenario sketched in \cref{ss:problem:interactive-teaching,fig:interaction}.
An example with a real robot 
is shown in \cref{fig:experiment-real-robot}.

\newcommand{\stepsubsection}[1]{\emph{#1:}\hspace{.5ex}}

\stepsubsection{1. Command}
Following 
\citep{Kartmann2021SemanticSceneManipulation},
we use a Named Entity Recognition (NER) model to parse object and relation phrases from a language command 
and ground them to objects and relations in the robot's memory using substring matching.
In addition, the resulting task $\trgrelationfull$ is stored in the relational command segment of the memory system. 
This is required to later construct the sample from a demonstration.

\stepsubsection{2. Plan}
We query the current model $\genmodel_{\trgsrel}^\timek$ from the spatial relation memory segment.
If $\genmodel_{\trgsrel}^\timek = \emptyset$, the geometric meaning of the spatial relation is unknown, and the robot cannot solve the task. 
In this case, the robot verbally expresses its lack of knowledge and requests a demonstration of what to do from the human.
Otherwise, $\genmodel_{\trgsrel}^\timek = \theta$ defines a cylindrical distribution, which is used to sample a given number of candidates (\num{50}~in our experiments).
As in \citep{Kartmann2021SemanticSceneManipulation}, 
the candidates are filtered for feasibility, 
ranked according to
the distribution's probability density function
$\func{p_\cdParam}{\cRadius, \cAzim, \cHeight}$,
and the best candidate is selected for execution.
We consider a candidate to be feasible if it is free of collisions, reachable and stable.
To decide whether a candidate is free of collisions,
the target object is placed virtually at the candidate position and tested for collisions with other objects
within a margin of \SI{25}{\milli\meter}
at \num{8} different rotations around the gravity vector
in order to cope with imprecision of action execution, \eg, the object rotating inside the robot's hand while grasping or placing
(note that our method aims to keep the object's orientation; the different rotations are only used for collision checking).
If a feasible candidate is found, planning is successful.
However, after filtering, 
it is possible that none of the sampled candidates is feasible.
This is especially likely when 
only few samples have been collected so far and, thus, 
the model's variance is low.
Before, this event marked a failure,
but in this work, the robot is able to recover by expressing its inability to solve the task and ask the human for a demonstration.
Note that the query mechanism of both failure cases, 
\ie~a missing model and an insufficient one, are structurally identical. 
In both cases, the human is asked to perform the task it originally gave to the robot, \ie, manipulate the scene to fulfill the relation~$\trgrelation$.

\stepsubsection{3a. Execution by Robot and 3b. Query}
If planning was successful~(3a.), the robot executes the task 
by grasping the target object and executing a placing action parameterized based on the selected placing position.
If planning failed~(3b.), the robot verbally requests a demonstration of the task from the human using sentences generated from templates and its speech-to-text system.
Examples of generated verbal queries are 
``I am sorry, I don't know what `right' means yet, can you show me what to do?'' (no model)
and 
``Sorry, I cannot do it with my current knowledge. Can you show me what I should do?'' (insufficient model).
Then, the robot waits for the speech cue signaling the finished demonstration.

\stepsubsection{4. Execution by Human and 5. Speech Cue}
After being queried for a demonstration (3b.), the human relocates the target object to fulfill the requested spatial relation.
To signal the completed demonstration,
the human gives a speech cue such as ``Place it here'' to the robot, 
that is detected using simple keyword spotting
and triggers the recording of a new sample.
This represents a third case where demonstrations can be triggered:
The robot may have successfully executed the task in a qualitative sense (3a.), but the human may not be satisfied with the chosen placing position.
In this case, the human can simply change the scene in order to demonstrate what the robot \emph{should have done}, and give a similar speech cue ``No, put it here.'' 
Again, note that this case is inherently handled by the framework without a special case: 
When the speech cue is received, the robot has all the knowledge it requires to create a new sample, 
independently of whether it executed the task before or  explicitly asked for a demonstration.

\stepsubsection{6. Observation and Model Update}
When the robot receives the speech cue,
it assembles a new sample by querying its memory for the relevant information.
It first queries its relational command segment for the latest command, which specifies the requested relation~$\trgrelation$ that was just fulfilled by the demonstration.
It then queries its object instance segment for the state of the scene $\poseset^\timek$ when the command was given and the current state $\poseset^{\timekx[k+1]}$.
Combined, this information forms a new sample
$\sample =\mtup{ \poseset^\timek, \trgrelation, \poseset^\trg }$.
Afterwards, the robot stores the new sample in its long-term memory
and updates $\genmodel_{\trgsrel}$ as described in \cref{ss:problem:interactive-teaching}.
Finally, the robot thanks the human demonstrator for the help, \eg, ``Thanks, I think I now know the meaning of `right' a bit better.''

\section{Results and Discussion}
\label{s:evaluation}

We evaluate our method 
quantitatively by simulating the interaction scheme with a virtual robot
and
(\cite{Asfour2019}).

\subsection{Quantitative Evaluation}

With our experiments, we aim at investigating two questions: 
(1) How many demonstrations are necessary to obtain a useful model for a given spatial relation?
And
(2) how does our model perform compared to baseline models using fixed offsets instead of learned cylindrical distributions?
To this end, 
we implement the proposed human-robot interaction scheme
based on human demonstrations collected in simulation.

\subsubsection{Experimental Setup}

\newcommand{\task}{T}
\newcommand{\taskdef}{\mtup{ \poseset^\timek, \rel }}
\newcommand{\taskfull}{\task = \taskdef}
\newcommand{\ntasks}{10}
\newcommand{\ntasksw}{{\ntasks}}  

For the design of our experiments,
a human instructs the robot to manipulate an object according to a spatial relation to other objects.
The robot tries to perform the task, and if it is successful, the interaction is finished.
If the robot is unable to solve the task,
it requests the user to demonstrate the task and updates its model of the spatial relation from the given demonstration.
We call this procedure one \emph{interaction}.
We are interested in how the robot's model of a spatial relation develops over the course of multiple interactions, 
where it only receives a new demonstration if it fails to perform the task at hand.

In our experiments,
we consider the \num{12} spatial relations $\srelset$ listed in
\cref{tab:eval:spatial-relations-and-baseline-models}.
Consider one relation $\srel \in \srelset$,
and assume we have \num{\ntasks} demonstrations of this relation.
As defined in \cref{eq:sample}, 
each demonstration $\sample_i$ $(i = 1, \dots, \ntasks)$ has the form
$\sample_i = \mtup{ \poseset^\timek_i, \rel_i, \poseset^\trg_i }$,
with initial object poses $\poseset^\timek_i$,
desired spatial relation~$\rel_i$,
and object poses after the demonstration $\poseset^\trg_i$.
Each demonstration $\sample_i$
also implicitly defines a \emph{task}~$\task_i = \mtup{ \poseset^\timek_i, \rel_i }$
consisting of only the initial scene and the desired relation.
Given this setup, we define a \emph{learning scenario} as follows:
First, we initialize a virtual robot that has no geometric model of $\srel$
as it has not received any demonstrations yet.
Then, we consecutively perform one interaction for each task $\task_1, \dots, \task_\ntasksw$.
After each interaction, we evaluate the robot's current model of $\srel$ on all tasks.

More precisely, 
in the beginning
we ask the robot to perform the first task~$\task_1$.
As it has no model of~$\srel$ at this point, it will always be unable to solve the task.
Following our interaction scheme,
the robot requests the corresponding demonstration 
and is given the target scene $ \poseset^\trg_i $.
The robot learns its first model of~$\srel$ from this demonstration,
which concludes the first interaction. 
We then evaluate the learned model on all tasks $\task_1, \dots, \task_\ntasksw$,
\ie, for each task we test whether the model can generate a feasible placing position for the target object.
Here, we consider $\task_1$ as \emph{seen}, and the remaining tasks $\task_2, \dots, \task_\ntasksw$ as \emph{unseen}.
Accordingly, we report the success ratios, \ie, the proportion of solved tasks, among the seen task, the unseen tasks and all tasks.
We then proceed with the second interaction, 
where the robot is asked to perform $\task_2$.
This time, the model learned from the previous interaction may already be sufficient to solve the task.
If this is the case, the robot does \emph{not} receive the corresponding demonstration
and thus keeps its current model.
Otherwise, it is given the new demonstration $\sample_2$ and can incrementally update its model of $\srel$.
Again, we test the new model on all tasks, 
where $\task_1$ and $\task_2$ are now considered as seen and $\task_3, \dots, \task_\ntasksw$ as unseen.
We continue in this manner for all tasks left.
After the last interaction, all tasks $\task_1, \dots, \task_\ntasksw$ are seen and there are no unseen tasks.
For completeness,
we also perform an evaluation on all tasks before the first interaction;
here, the robot trivially fails on all tasks since it initially has no model of~$\srel$.

Overall, for one learning scenario of relation $\srel$,
we report the proportions of solved tasks among all tasks, the seen tasks and the unseen tasks after each interaction (and before the first interaction).
Note that, as explained above, the number of seen and unseen tasks change over the course of a learning scenario.
In addition, we report the number of demonstrations the robot has been given after each interaction.
Note that this number may be smaller than the number of performed interactions, 
as a demonstration is only given during an interaction if the robot fails to perform the task at hand.
As the results depend on the order of the tasks and demonstrations, 
we run \num{10} repetitions of each learning scenario with the tasks randomly shuffled for each repetition,
and report all metrics' means and standard deviations over the repetitions.
Finally,
to obtain an overall result over all spatial relations,
we aggregate the results of all repetitions of learning scenarios of all relations $\srel \in \srelset$,
and report the resulting means and standard deviations 
of success ratios and number of demonstrations
with respect to the number of performed interactions.

\begin{table}
  \centering
  \def\arraystretch{1.4}
  \newcommand{\row}[4]{
    \emph{#2}
    & #3
    & #4
    \\
  }
  \newcommand{\dirvec}[1]{\func{\mathrm{dir}}{#1}}
  \begin{tabularx}{\linewidth}{c|Xc}
    \toprule
    \textbf{Spatial Relations}
    &
    \multicolumn{2}{c}{\textbf{Baseline Model}}
    \\
    
    $(\srel \in \srelset)$
    &
    \textbf{Description}
    &     
    \textbf{Placing Position} $\pos_\reltrg^\trg$
    
    \\ \midrule
    
    \row{1}{right of}{
      \SI{20}{\centi\meter} in $+x$ direction
    }{
      $\pos_\relref + \SI{20}{\centi\meter}\cdot \mvec{x}$
    }
    
    \row{2}{left of}{
      \SI{20}{\centi\meter} in $-x$ direction
    }{
      $\pos_\relref - \SI{20}{\centi\meter}\cdot \mvec{x}$
    }
    
    \row{3}{behind}{
      \SI{20}{\centi\meter} in $+y$ direction
    }{
      $\pos_\relref + \SI{20}{\centi\meter}\cdot \mvec{y}$
    }
    
    \row{4}{in front of}{
      \SI{20}{\centi\meter} in $-y$ direction
    }{
      $\pos_\relref - \SI{20}{\centi\meter}\cdot \mvec{y}$
    }
    
    \row{5}{on top of}{
      \SI{10}{\centi\meter} in $+z$ direction
    }{
      $\pos_\relref + \SI{10}{\centi\meter}\cdot \mvec{z}$
    }
    
    \row{6}{close to}{
      \SI{20}{\centi\meter} towards initial position of $\reltrg$
    }{
      $\pos_{\relref} + \SI{10}{\centi\meter} \cdot \dirvec{\pos_{\reltrg} - \pos_{\relref}}$
    }
    
    \row{7}{far from}{
      \SI{60}{\centi\meter} towards initial position of $\reltrg$
    }{
      $\pos_{\relref} + \SI{50}{\centi\meter} \cdot \dirvec{\pos_{\reltrg} - \pos_{\relref}}$
    }
    
    \row{8}{between}{
      Mid point of reference objects
    }{
      $\nicefrac{1}{n}\cdot\sum_{i=1}^{n}\pos_{\relref_i}$
    }
    
    \row{9}{among\rule[-4.5mm]{0pt}{11mm}%
    }{
      \makecell[l]{
        \SI{20}{\centi\meter} from reference objects' \\ 
        mid point towards initial position of $\reltrg$
      }
    }{
      $\pos_{\relref} + \SI{10}{\centi\meter} \cdot \dirvec{\pos_{\reltrg} - \nicefrac{1}{n}\cdot\sum_{i=1}^{n}\pos_{\relref_i}}$
    }
    
    \row{10}{closer}{
      Half the distance from reference to $\reltrg$
    }{
      $ \pos_{\relref}
      + \nicefrac{1}{2} \cdot \left(\pos_{\reltrg} - \pos_{\relref}\right)$
    }
    \row{11}{farther from}{
      Twice the distance from reference to $\reltrg$
    }{
      $ \pos_{\relref}
      + 2 \cdot \left(\pos_{\reltrg} - \pos_{\relref}\right)$
    }
    \row{12}{on the other side of%
      \rule[0mm]{0pt}{6.5mm}%
    }{
      \makecell[l]{
        Distance between reference and target \\
        in opposite direction of target%
      }%
    }{
      $ \pos_{\relref}
      -	\left(\pos_{\reltrg} - 	\pos_{\relref}\right)$
    }
    
    \bottomrule
  \end{tabularx}
  \caption{
    Left column:
    Spatial relation symbols $\srel \in \srelset$
    used in the experiments.
    Right columns:
    Their corresponding baseline model,
    where
    $\pos_\relref, \pos_{\relref_i} \in \reals^3$
    refer to the positions of reference objects,
    $\pos_\reltrg$~refers to the initial position of target object $\reltrg$,
    $\mvec{x}, \mvec{y}, \mvec{z}$ are the unit vectors
    in the robot's coordinate system,
    and
    \mbox{$\dirvec{\mvec{v}} = \nicefrac{\mvec{v}}{\left\| \mvec{v} \right\|}$}
    is the direction of a vector $\mvec{v} \in \reals^3$.
  }
  \label{tab:eval:spatial-relations-and-baseline-models}
\end{table}

We compare our method with baseline models for all relations
which place the target object at a fixed offset from the reference objects.
\cref{tab:eval:spatial-relations-and-baseline-models}
gives their descriptions 
and the definitions of the placing position $\pos_\reltrg^\trg$ they generate.
The baseline models of direction-based static relations 
(\emph{right of}, \emph{left of}, \emph{behind}, \emph{in front of}, \emph{on top of})
yield a candidate at a fixed distance towards the respective direction.
Distance-based static relations 
(\emph{close to}, \emph{far from}, \emph{among})
place the target object at a fixed distance towards its initial direction.
The baseline models of dynamic spatial relations 
(\emph{closer}, \emph{farther from}, \emph{on the other side})
are similar, but the distance of the placing position to the reference object is defined relative to the current distance between the objects.
We conduct the experiments on the baseline models in the same way as described above,
except that
(1) the robot has access to the models from the beginning,
and 
(2)~we only report the success ratio among all tasks over one repetition, 
as the baseline models are constant and do not change during a learning scenario or depending on the order of tasks.

For collecting the required demonstrations for each relation,
we use a setup with a human, several objects on a table in a simulated environment
and a command generation based on sentence templates.
Given the objects present in the initial scene 
and the set of spatial relation symbols~$\srelset$,
we generate a verbal command specifying a desired spatial relation
$\relationfull$ 
such as ``Place the cup between the milk and the plate.''
The human is given the generated command
and performs the task by moving the objects in the scene.
We record 
the initial scene~$\poseset^\timek$,
the desired spatial relation~$\rel$,
and the final scene $\poseset^\trg$,
describing a full demonstration
$\sample = \mtup{ \poseset^\timek, \rel, \poseset^\trg }$
as required for the experiment.
Aside from the objects involved in the spatial relation,
the scenes also contain inactive objects which can lead to collisions when placing the object at a generated position,
thus rendering some placements infeasible.

\subsubsection{Results}

\begin{figure}[tb]
  \centering
  \includegraphics[width=.9\linewidth]{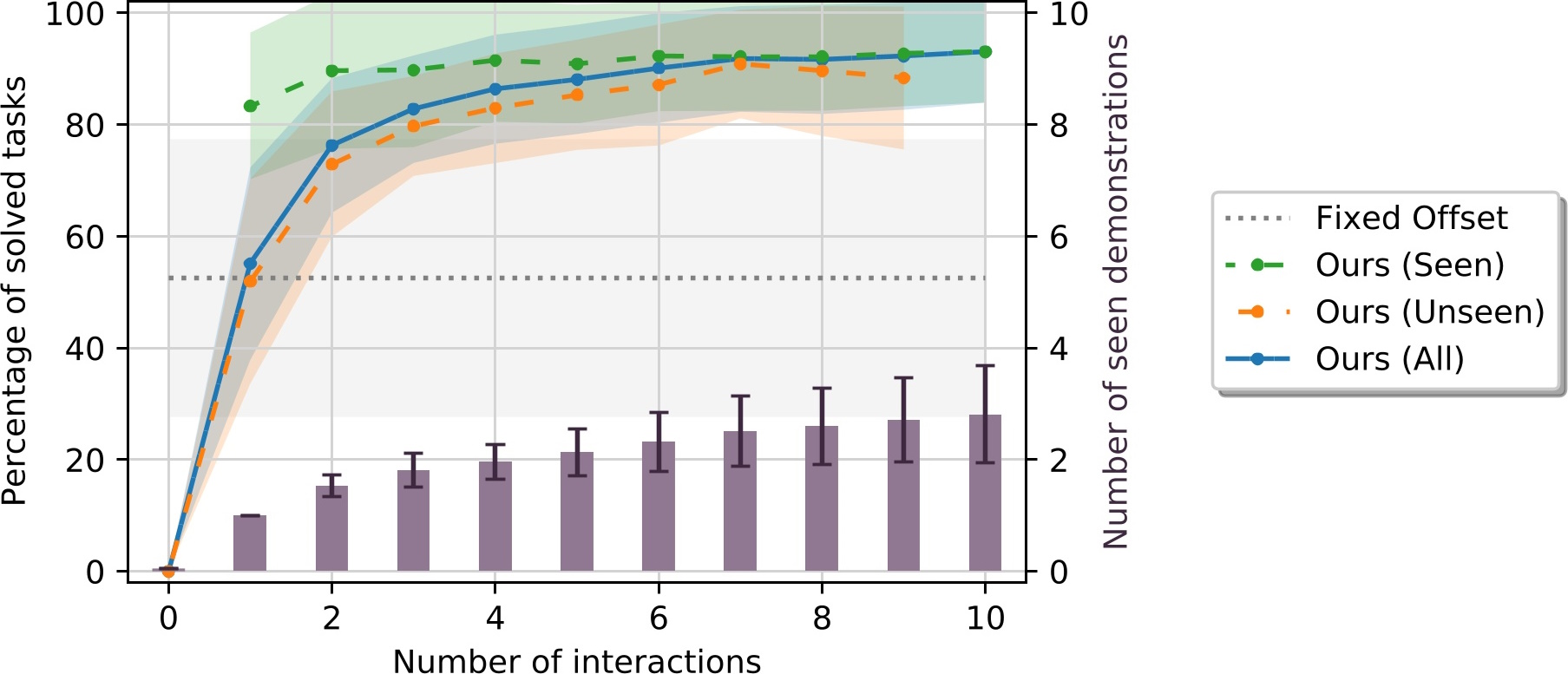}
  \caption{
    Means and standard deviations of
    percentage of tasks solved
    by our method
    among the 
    \textcolor[HTML]{2ca02c}{seen (green)}, 
    \textcolor[HTML]{ff7f0e}{unseen (orange)} and 
    \textcolor[HTML]{1f77b4}{all (blue)} 
    tasks,
    by the 
    \textcolor[HTML]{808080}{baseline models using fixed offsets (gray)}
    as well as 
    \textcolor[HTML]{3d283f}{number of demonstrations (purple)} the robot has been given
    after a given number of interactions.
    All metrics are
    aggregated over multiple repetitions and all spatial relations.
  }
  \label{fig:evaluation:quantitative-all}
\end{figure}

\cref{fig:evaluation:quantitative-all} 
shows
the means and standard deviations
of the percentage of solved tasks
among all tasks, the seen tasks and the unseen tasks
after each interaction
aggregated over all relations and repetitions
as described above.
Furthermore,
it shows the averages of total number of demonstrations the robot has received
after each interaction.
As explained above, 
note that not all interactions result in a demonstration.
Also, note that the number of seen and unseen tasks change with the number of interactions;
especially,
there are no seen tasks before the first interaction
and there are no unseen tasks after the last one.
For comparison,
the success ratio of the baseline models averaged over all relations
are shown as well.

\begin{figure}[tb]
  \centering
  \includegraphics[width=\linewidth]{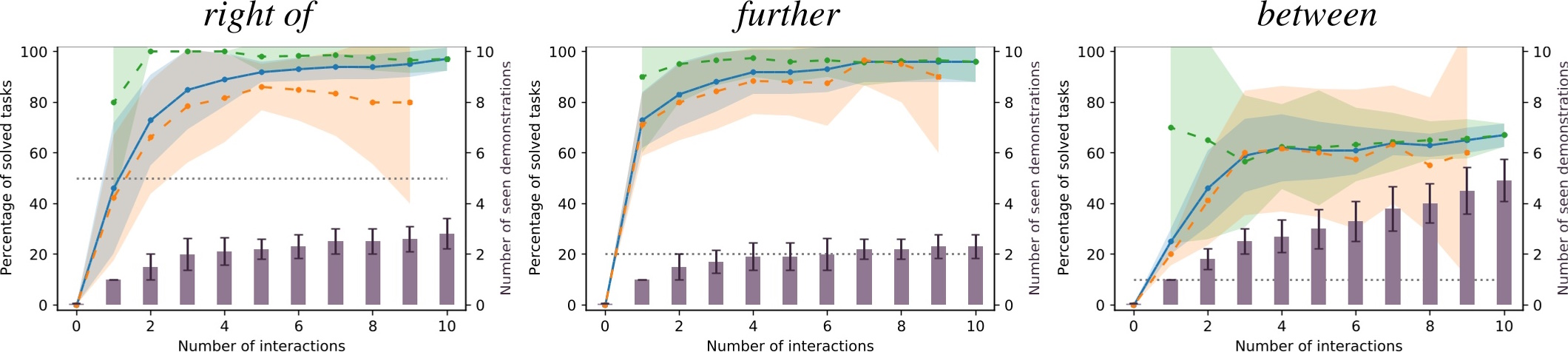}
  \caption{
    Examples of success ratios and number of demonstrations 
    aggregated over the repetitions of single relations.
    Colors are as in \cref{fig:evaluation:quantitative-all}.
    Note that for single relations, the success ratios of the baseline models are constant 
    and, thus, have no variance.
  }
  \label{fig:evaluation:quantitative-singles}
\end{figure}

\begin{figure}[tb]
  \centering
  \includegraphics[width=\linewidth]{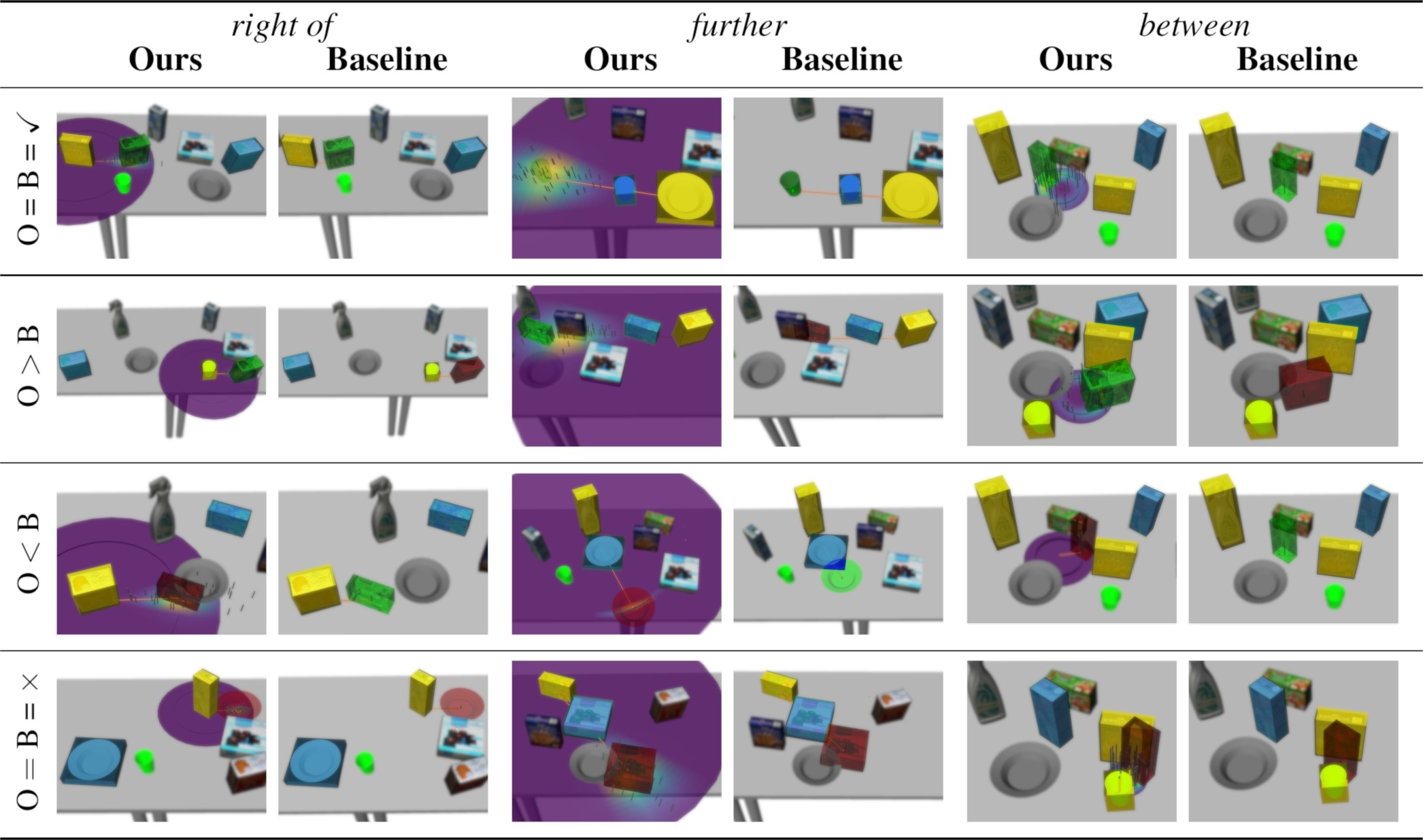}
  \caption{
    Examples of solved and failed tasks from our experiment.
    In all images, the reference objects are highlighted in yellow,
    the target object is highlighted in light blue,
    and the generated placing position is visualized in green if successful and red if no feasible candidate was found.
   	The orange arrow indicates the mean of a cylindrical distribution (Ours) and the placement according to the baseline model, respectively.
   	The cylindrical distribution's p.d.f. is visualized around the reference objects (low values in purple, high values in yellow).
   	Sampled positions are shown as grey marks.
    The left column of each relations shows our model, 
    while the right column shows the result of the baseline model in the same task.
    For our model, the current cylindrical distribution is visualized as well.
    For good qualitative coverage,
    each row show another case with respect to the models' success:
    (1) both succeed,
    (2) ours succeeds, baseline fails,
    (3) ours fails, baseline succeeds,
    (4) both fail.
  }
  \label{fig:evaluation:quantitative-singles-examples}
\end{figure}

In addition,
\cref{fig:evaluation:quantitative-singles}
gives examples of success ratios and number of received demonstrations 
aggregated only over the repetitions of learning scenarios of single spatial relations.
Note that here, the baseline's performance is constant over all repetitions as the number of tasks it can solve is independent of the order of tasks in the learning scenario.
Finally,
\cref{fig:evaluation:quantitative-singles-examples}
presents concrete examples of solved and failed tasks involving the spatial relations in \cref{fig:evaluation:quantitative-singles}
demonstrating the behavior of our method and the baseline 
during the experiment.
Note that the cases represented by the rows in 
\cref{fig:evaluation:quantitative-singles-examples}
are not equally frequent; 
our method more often outperforms the baseline models than vice-versa
as indicated by the success ratios in 
\cref{fig:evaluation:quantitative-all,fig:evaluation:quantitative-singles}.

\newcommand{\evalOursLtBaseline}{O $<$ B}
\newcommand{\evalOursEqBaselineEqFail}{O $=$ B $=$ $\times$}

\subsubsection{Discussion}

The baseline models achieved only \SI{52.5 \pm 24.9}{\percent} success rate over all relations,
showing that many tasks in our demonstrations
were not trivially solvable using fixed offsets.
Also, the standard deviation of 52.5 $\pm$ 24.9 \%.
The models were successful if their candidate placing position was free and on the tables.
However, if the single candidate was,
\eg, obstructed by other objects,
the models could not fall back to other options.
This was especially frequent among relations such as 
\emph{close to}, \emph{between} and \emph{among}
which tend to bring the target object close to other objects.
Other collisions were caused because the sizes of the involved objects were not taken into account, 
especially among relations with fixed distances.

As expected, using our method the robot could never solve a task before the first interaction.
However, after the first interaction (which always results in a demonstration),
the robot could already solve
about~\SI{83.3 \pm 13.1}{\percent} of seen and
about~\SI{51.9 \pm 18.3}{\percent} of unseen tasks
on average,
almost equaling the baseline models on the unseen tasks.
After just two interactions, the mean success ratios on seen tasks reaches a plateau of
at \SIrange{89}{92}{\percent}, 
with \num{1.53 \pm 0.19} demonstrations on average.
Importantly, the success ratios among the seen tasks
stays high and does not decrease after more interactions,
which indicates that our method generally does not ``forget'' what it has learned from past demonstrations.
The success ratios among unseen tasks rise consistently with the number of interactions,
although more slowly than among seen tasks.
Nonetheless, after five interactions,
the robot could solve 
\SI{85.3 \pm 9.9}{\percent} of unseen tasks
after having received 
\num{2.13 \pm 0.41} demonstrations on average,
which shows that our method can generalize from few demonstrations.
After completing each learning scenario,
\ie, after all interactions have been performed,
the robot could solve 
\SI{93.0 \pm 09.1}{\percent}
of all tasks
while having received 
\num{2.81 \pm 0.87} demonstrations on average.

One might wonder why the robot is not always able to 
successfully reproduce the first demonstration on the single seen task after the first interaction.
This can be explained in two ways:
First, 
the human is free to change the target object's orientation during the demonstration,
which can allow the human to place it closer to other objects than without rotating it.
However, as our method only generates new positions while trying to keep the original object orientation,
the robot cannot reproduce this demonstration as it would lead to a collision.
Second,
we use a rather conservative strategy for collision detection in order to anticipate inaccuracies in action execution (\eg, the object rotating in the robot's hand during grasping or placing).
More precisely,
we use a collision margin of \SI{25}{\milli\meter} to
check for collisions at different hypothetical rotations of the target object (see \cref{ss:approach:interactive-teaching}).
Therefore, 
if the human placed the object very closely to another in the demonstration,
the robot might discard this solution to avoid collisions.
These are situations that can only be solved after increasing the model's variance through multiple demonstrations.

We can observe the behavior of our method in more detail by focusing on the results of single relations shown in \cref{fig:evaluation:quantitative-singles} and the examples in \cref{fig:evaluation:quantitative-singles-examples}.
First, 
the standard deviation over the success ratios among the unseen tasks tends to increase towards the end of the learning scenarios.
This is likely due to the decreasing number of unseen tasks towards the end:
Before the final interaction, there is only one unseen task left, so the success ratio is either
\SI{0}{\percent} or \SI{100}{\percent} in each repetition, 
leading to a higher variance than when the success ratio is averaged over more tasks.
As for the relation \emph{right of},
common failure cases were caused by the conservative collision checking in combination with a finite number of sampled candidates 
(\cref{fig:evaluation:quantitative-singles-examples}, third row), 
or the mean distance of the learned distribution being too small for larger objects such as the plate 
(\cref{fig:evaluation:quantitative-singles-examples}, fourth row).
With the relation \emph{further},
failures were often caused by the candidate positions being partly off the table
in combination with the distance variance being too small to generate alternatives
(\cref{fig:evaluation:quantitative-singles-examples}, third row),
or scenes where the sampled area was either blocked by other objects or off the table
(\cref{fig:evaluation:quantitative-singles-examples}, fourth row).

The relation \emph{between} was one of the relations that were more difficult to learn,
with a success ratio of only \SI{67.0 \pm 04.6}{\percent} among all tasks 
after receiving an average of \num{4.90 \pm 0.83} demonstrations 
at the end of the learning scenario
(the baseline model achieved only \SI{10}{\percent}).
In the demonstrations,
the area between the two reference objects was
often cluttered, which prevented our method from finding 
a collision free placing location.
The success ratio among the seen tasks
starts at \SI{70.0 \pm 45.8}{\percent}
after one interaction.
The large variance indicates that,
compared to other relations,
there were many demonstrations that could not be reproduced after the first interaction,
with the success ratio among seen tasks being either
\SI{0}{\percent} or \SI{100}{\percent}
leading to a high variance, 
similar to the success ratios among unseen tasks towards the end of the learning scenarios.
Moreover, 
the success ratio among the seen tasks
decreases to \SI{56.7 \pm 26.0}{\percent}
after the third interaction,
although it slightly increases again afterwards.
In this case, 
the \num{1.50 \pm 0.50} additional demonstrations
caused the model to ``unlearn'' to solve the first tasks
in some cases.
However, after the third interaction, the success ratios among seen and unseen tasks stabilize
and do not change significantly with more demonstrations.
Apparently, the models reached their maximum variance after a few interactions,
with new demonstrations not changing the model significantly;
however our conservative collision detection often caused all candidates to be considered infeasible.
One especially difficult task is shown in
\cref{fig:evaluation:quantitative-singles-examples} (fourth row),
where the two reference objects were standing very close to each other, leaving little space for the target object.
Finally, note that the tasks for the \emph{between} relation 
shown in the first and the third row of \cref{fig:evaluation:quantitative-singles-examples} are the same.
This is because the baseline could only solve this single task.
The corresponding failure example of our model (third row)
was shows a model learned from only one demonstration.
Indeed, with two or more demonstrations, 
our method was always able to solve the this task
(example in first row).

To summarize,
while some aspects can still be improved,
the overall results demonstrate
that our generative models of spatial relations
can be effectively learned in an incremental manner from few demonstrations,
while our interaction scheme allows 
the robot to obtain new demonstrations
to improve its models if they prove insufficient for a task at hand.

\subsection{Validation Experiments on Real Robot}
\label{ss:evaluation:real-robot}

We performed validation experiments on the real humanoid robot (\cite{Asfour2019}) which are shown in the video\footnote{\url{https://youtu.be/x6XKUZd_SeE}}.
An example is shown in~\cref{fig:experiment-real-robot}{} and described in more detail here:
In the beginning, the robot has no geometric model of the \emph{right of} relation.
First, the user commands the robot to place an object \emph{on the right side of} a second object (step 1. in \cref{ss:problem:interactive-teaching}).
The robot grounds the relation phrase ``on the right side of'' to the respective entry in its memory (2.), 
and responds that it has ``not learned what \emph{right} means yet,'' and requests the user to show it what to do (3a.).
Consequently, the user gives a demonstration by performing the requested task (4.) and gives the speech cue ``Put it here'' (5.).
The robot observes the change in the scene and creates a model of the spatial relation \emph{right of} (6.).
Afterwards, the user starts a second interaction by
instructing the robot to put the object \emph{on the right side of} a third one (1.).
This time, the robot has geometric model of \emph{right of} in its memory (2.) and is able to perform the task (3b.).
Beyond that, we show examples of 
demonstrating and 
manipulation the scene according to 
the relations 
\emph{in front of}, \emph{on top of}, \emph{on the opposite side of}, and \emph{between}.

\section{Conclusion and Future Work}
\label{s:conclusion}

In this work, we described how a learning humanoid robot which has the task to manipulate the scene based on desired spatial object relations
can query and use demonstrations from a human during interaction
to incrementally learn generative models of spatial relations.
We demonstrated
how the robot can communicate its inability to solve a task 
in order to collect more demonstrations in a continual manner.
In addition, we showed
how a parametric representation of object spatial relations can be learned incrementally from few demonstrations.
In future work, 
we would like to make the human-robot interaction even more natural
by detecting when a demonstration is finished, 
thus releasing the requirement of a speech cue indicating this event.
Furthermore, we want to explore how knowledge about different spatial relations can be transferred between them and leveraged for learning new ones.

\section*{Conflict of Interest Statement}

The authors declare that the research was conducted in the absence of any commercial or financial relationships that could be construed as a potential conflict of interest.

\section*{Author Contributions}

RK developed the methods and their implementation and performed the evaluation experiments.
The entire work was conceptualized and supervised by TA.
The initial draft of the manuscript was written by RK and revised jointly by RK and TA.

\section*{Funding}

\ackJuBotOML

\section*{Supplemental Material}

This paper is supplemented by a video 
giving an overview of our approach
and showing the validation experiments described in \cref{ss:evaluation:real-robot}.

\section*{Data Availability Statement}

The raw data supporting the conclusions of this article will be made available by the authors, without undue reservation.

\bibliographystyle{Frontiers-Harvard} 
\bibliography{references}



\end{document}